\documentclass[letterpaper, 10 pt, conference]{ieeeconf} 

\IEEEoverridecommandlockouts

\usepackage{graphics} 
\usepackage{epsfig} 
\usepackage{mathptmx} 
\usepackage{times} 
\usepackage{amsmath} 
\usepackage{amssymb}
\usepackage{xcolor}

\newtheorem{Th}{Theorem}
\newtheorem{Remark}{Remark}
\newtheorem{assump}{Assumption}

\newcommand{\pos}{\textbf{r}}
\newcommand{\C}{\textbf{c}}

\graphicspath{{./images/}}

\title{\LARGE \bf
Voronoi-based Multi-Robot Formations for 3D Source Seeking \\ via Cooperative Gradient Estimation
}

\author{Lara~Bri\~{n}\'{o}n-Arranz$^{1}$, Martin Abou Hamad$^{1}$ and Alessandro Renzaglia$^{2}$
\thanks{$^{1}$Univ. Grenoble Alpes, CNRS, Grenoble INP, GIPSA-lab, 38000 Grenoble, France. Email: {\tt\small firstname.lastname@grenoble-inp.fr}}
\thanks{$^{2}$Inria, INSA Lyon, CITI, UR3720, 69621 Villeurbanne, France. Email: {\tt\small firstname.lastname@inria.fr}}
}

\begin{document}

\maketitle
\thispagestyle{empty}
\pagestyle{empty}

\begin{abstract}
    In this paper, we tackle the problem of localizing the source of a three-dimensional signal field with a team of mobile robots able to collect noisy measurements of its strength and share information with each other. The adopted strategy is to cooperatively compute a closed-form estimation of the gradient of the signal field that is then employed to steer the multi-robot system toward the source location. In order to guarantee an accurate and robust gradient estimation, the robots are placed on the surface of a sphere of fixed radius. More specifically, their positions correspond to the generators of a constrained Centroidal Voronoi partition on the spherical surface. We show that, by keeping these specific formations, both crucial geometric properties and a high level of field coverage are simultaneously achieved and that they allow estimating the gradient via simple analytic expressions. We finally provide simulation results to evaluate the performance of the proposed approach, considering both noise-free and noisy measurements. In particular, a comparative analysis shows how its higher robustness against faulty measurements outperforms an alternative state-of-the-art solution.
\end{abstract}

\section{Introduction}
\label{sec:intro}

Cooperative source seeking is a fundamental problem in multi-robot applications because of its relevance in numerous different scenarios such as search and rescue operations, environmental monitoring, and pollution source detection among others \cite{bayat2017survey}. In this problem, a spatially distributed signal field is generated by a source of unknown location. The team of autonomous robots is equipped with sensors able to measure the signal intensity at their locations. The final goal is then to cooperatively localize the source of the signal by exploiting the information gathered by each robot.

\begin{figure}
    \centering
    \includegraphics[width=0.85\linewidth]{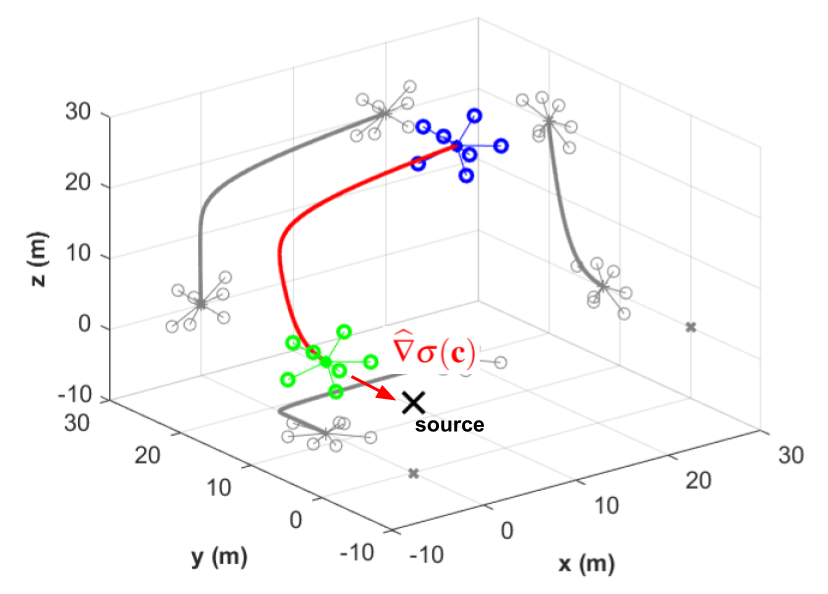}
    \caption{Cooperative source seeking in 3D: a formation of $7$ robots follows the estimated gradient of a signal field to converge toward its source.}
    \label{fig:source-seeking}
\end{figure}

Numerous solutions have been proposed over the years to deal with the source-seeking problem with both single and multi-robot systems, but mostly limited to 2D environments. Some works presented gradient-free approaches such as the Speeding-Up and Slowing-Down method inspired by schools of fish seeking darker areas \cite{Wu2015}, the consensus-like Principal Component Analysis perception algorithm without explicit communication \cite{al2018gradient}, and Particle Swarm Optimization based algorithms \cite{zou2015particle}, \cite{zhang2023pso}. However, the vast majority of the proposed strategies compute a local estimation of the signal gradient based on the collected measurements and use it to drive the robots toward the source location, as illustrated in Fig.~\ref{fig:source-seeking}. Obtaining an accurate estimation from sparse and usually noisy data remains nevertheless a challenging problem to solve. Moreover, only very few works addressed the problem in 3D environments, which better represent many real-world scenarios. Possible approaches include consensus-based algorithms \cite{fabbiano2014source} and cooperative Kalman filters \cite{Wu2011}. However, these strategies involve high communication and computational complexities, which are crucial aspects in the presence of small limited robotic platforms, and do not explicitly quantify the quality of the estimation algorithm, nor analyze the impact of measurement noise. Furthermore, in \cite{fabbiano2014source}, the 3D formation can only be defined for specific numbers of robots, such as $N=\{6,12,20,32,...\} $.

To ensure formal guarantees on the obtained estimation of the gradient, an alternative interesting solution is to define multi-robot formations or configurations respecting specifically defined geometric constraints. Also in this case, the majority of works consider exclusively 2D environments, as in \cite{marzat2015cooperative, renzaglia2020search}.
Within this line, in our previous work \cite{Lara2019TRO}, we showed that by adopting particular symmetric multi-robot formations, in both two and three dimensions, the estimates of the gradient and the Hessian matrix of an unknown signal can be obtained via simple averages that involve only the products of the measurements collected by the robots and their relative position with respect to the formation center. 
Despite the advantages brought by these formations, this approach presents a few drawbacks. Firstly, it requires to have an even number of robots. This may raise some problems in several scenarios, for instance in the case of a defective robot which would force the formation to reconfigure with the remaining $N-1$ robots. Furthermore, each robot needs to know its specific role within the formation to compute its position, requiring a centralized architecture. Finally, the proposed 3D formations are composed of two circles lying on two parallel planes that limit the spatial dispersion of the available sensors. This may produce some redundancy in the gathered data, especially with large teams.

The main contribution of this paper is to present a new multi-robot formation strategy able to overcome the limitations of the current state of the art and robustly estimate the gradient of a 3D scalar field to localize its source. In particular, we propose and analyze multi-robot formations that correspond to Centroidal Voronoi partitions of spherical surfaces. The advantages of this novel approach are:
\begin{itemize}
    \item Suitable multi-robot formations can be defined for any number of robots $N$, with $N \geq 4$.
    \item The formations can be achieved in a distributed way, easing reconfiguration after possible robots' failures.
    \item Robots are spatially uniformly distributed resulting in a robust gradient estimation of the signal, that can be obtained via simple algebraic computations and whose error remains upper bounded.
\end{itemize}

\section{Problem Formulation \& Preliminaries}
\label{sec:problem}

The main objective of the cooperative source-seeking problem is to guide a team of robots to navigate through a signal field and localize its source. In this scenario, each robot represents a mobile sensor able to collect measurements of the signal strength emitted by the source.
More formally, the signal distribution $\sigma(\pos):\mathbb{R}^3 \rightarrow \mathbb{R}^+$ is a function representing the scalar field at location $\pos$ achieving its maximum at position $\pos^*$ where the source is located and smoothly decreasing to zero far from the source. Denoting the signal gradient at a location $\pos$ as $\nabla\sigma(\pos)$, we consider the following assumption on the signal:

\begin{assump}\label{assum:signal3D}
The function $\sigma : \mathbb{R}^3 \rightarrow \mathbb{R}^+$ is two times continuously differentiable, i.e., $\sigma \in \mathcal{C}^2$, and all its partial derivatives up to order two are globally bounded. Moreover, $\nabla\sigma(\pos^*) = 0$ and $\nabla\sigma(\pos) \neq 0, \, \forall \pos \neq \pos^*$. This implies that there exists a scalar $L$ such that
$$
|\varphi(\pos,\C)|:=| \sigma(\pos) - \sigma(\C)-\nabla\sigma(\C)^T (\pos - \C)| \leq L \|\pos - \C\|^2
$$
where $|\cdot|$ denotes the absolute value, $\|\cdot\|$ denotes the Euclidean norm, and $\varphi (\pos, \C)$ corresponds to the first-order remainder of the Taylor expansion about the point $\C$.
\end{assump}

The gradient $\nabla\sigma$ can then be used to drive the sensor network towards the signal source via gradient-ascent (GA) methods by steering the center of the formation as follows:
$$
\C(k+1)=\C(k)+\epsilon\nabla\sigma(\C(k))
$$
where $\C(k)\in\mathbb{R}^3$ represents the formation center at iteration $k$ and $\epsilon > 0$ is the step size of the GA algorithm.
However, the gradient is usually not directly available and the objective is thus to design a multi-robot strategy able to cooperatively compute a gradient estimate, denoted by $\widehat \nabla\sigma(\pos)$, based on the collected measurements of the signal strength. 

Consider a system composed of $N$ robots whose positions in the inertial global frame are denoted by $\pos_i \in \mathbb{R}^3$ with $i=1, \ldots, N$. The robots are equipped with the appropriate sensors to measure the signal field of interest. Let $\sigma(\pos_i)$ denote the measurement of the signal strength collected by robot $i$. The robots can be controlled to stabilize their states, position, and velocity, to a desired reference. As presented in a large number of works dealing with robot control, the non-linear dynamics of the robots can be linearized and/or simplified in order to apply several commonly used control techniques (see \cite{Nascimento2019} for an exhaustive survey) and ensure the tracking of a desired position.
In this paper, the desired position $\pos_i^{ref}$ for each robot $i$ in the team is provided by the multi-robot formation generator. The study of the low-level control of the robots is out of the scope of this work, consequently, we assume that the position controller of each robot guarantees a good tracking of this reference position, such that $\pos_i \rightarrow \pos_i^{ref}$. The robots are then controlled to maintain the desired formation while moving following the formation center. To consider realistic velocity constraints for the robots, we impose that the formation center moves with a limited velocity, i.e., $\|\C(k+1)-\C(k)\| \leq \gamma$ where $\gamma>0$ is a control parameter to be designed taking into account the physical limitations of the robots.

The solution proposed in this paper exploits some of the findings published in our previous work on symmetric formations for gradient estimation \cite{Lara2019TRO}. For the sake of clarity, we here briefly recall the necessary results.

Consider an even number of robots $N\!=\!2n, \, n\!\in\!\mathbb{N}$ forming a symmetric configuration composed of two parallel circular formations whose centers are aligned with the $z-$axis. The center point $\C$ is located between the two circles at distance $D\sin{\theta_F}$ from each one. One half of the robots is uniformly distributed in the upper circular formation, the other half is uniformly distributed in the lower circular formation and thus, their relative vectors with respect to the center point $\C$ are also evenly spaced. The robots' positions can be expressed in spherical coordinates as follows:
\begin{equation}\label{eq:Symmetric_formation}
\begin{split}
\pos_i&=\C+D \left[\sin{\theta_i} \cos{\phi_i},
\, \sin{\theta_i} \sin{\phi_i}, \, \cos{\theta_i}\right]^T\\
\theta_i&=\left\{\begin{array}{ll}
   \theta_F, & \mbox{if} \quad i=2k-1\\
   \pi-\theta_F, & \mbox{if} \quad i=2k, 
\end{array}\right.
\quad k=1,\ldots,n
\end{split}
\end{equation}
where $\phi_i=2\pi i/N$ is the azimuthal angle, $D$ the radial distance to the center $\C$, $\theta_i$ is the polar angle and $\theta_F$ is defined such that $\sin{\theta_F}=\sqrt{2/3}$. 

This Symmetric Cylindrical formation has been proposed to satisfy three interesting and convenient properties. Firstly, all the robots are placed at the same distance from the center, i.e., $\|\pos_i-\C\|=D, \forall i$. Additionally, as proven in \cite{Lara2019TRO}, when $N=2n, \, n\geq 2$, the relative position vectors of the robots with respect to the center satisfy the following properties: 
\begin{equation}\label{eq:Symmetric_properties}
\sum_i^N (\pos_i-\C)= 0 \quad \mbox{and} \quad \sum_i^N (\pos_i-\C)(\pos_i-\C)^T =  \frac{ND^2}{3}\textbf{I}
\end{equation}
where $\textbf{I}\in \mathbb{R}^{3\times3}$ represents the identity matrix. Considering the Symmetric Cylindrical formation defined by \eqref{eq:Symmetric_formation}, the gradient of a 3D signal distribution $\sigma(\pos)$ can be estimated as presented in the following theorem:
\begin{Th}\label{th:Gradient_symmetric}
\textit{(from \cite{Lara2019TRO})} Assume that $\sigma(\pos):\mathbb{R}^3\rightarrow\mathbb{R}^+$ satisfies Assumption~\ref{assum:signal3D}. Considering a team of $N=2n$ robots with $n\geq 2$ forming a configuration given by \eqref{eq:Symmetric_formation} and defining
\vspace{-.0cm}
\begin{equation}\label{eq:Gradient_estimation}
\widehat \nabla \sigma(\C)=\frac{3}{ND^2}\sum_{i=1}^N\sigma(\pos_i)(\pos_i-\C)
\end{equation}
then it holds
\vspace{-.1cm}
\begin{equation}\label{Gradient_estimation_bound}
\| \widehat \nabla \sigma(\C)-\nabla \sigma(\C) \|\leq 3LD .
\end{equation}
\end{Th}
\vspace{.15cm}

The Symmetric Cylindrical formation proposed in \cite{Lara2019TRO} presents however some limitations: {\it 1)} an even number of robots is needed to satisfy the symmetric geometrical properties; {\it 2)} a central station is required to assign an identifier $i$ to each robot allowing defining their positions; {\it 3)} all the robots are located on two parallel circular formations, resulting in a limited distribution over the 3D space.

Inspired by this previous result, our objective is to propose a new formation that overcomes such limitations, i.e., allowing for any number of robots, avoiding the need for a central station, and achieving more dispersed measurements resulting in a higher coverage of the signal.

\section{Voronoi-based Multi-Robot Formation}
\label{sec:voronoi}

To achieve these objectives, we propose a new multi-robot formation strategy where the robots generate a Centroidal Voronoi Tessellation (CVT) of a spherical surface. As we show in this section, this solution allows the system to reach a high spatial dispersion while keeping important geometrical properties. Furthermore, each robot can now identify its position only based on the location of its teammates, i.e., without necessitating a centralized architecture.

Voronoi tessellation is a fundamental concept in Locational Optimization theory \cite{okabe1992} often used to design multi-robot strategies, especially for optimal coverage \cite{cortes2004coverage, renzaglia2020common, lorenzo2022coverage}.
Given a bounded set $\Omega \in \mathbb{R}^n$ and $N$ points
$\{\mathbf{p}_i\}_{i=1}^N$, we define the Voronoi tessellation of $\Omega$ generated by $\{\mathbf{p}_i\}_{i=1}^N$ as $\{V_i\}_{i=1}^N$:
$
 V_i = \{\mathbf{u}\in\Omega : ||\mathbf{u}-\mathbf{p}_i||<||\mathbf{u}-\mathbf{p}_j||\;\forall j\neq i\},\, i=1,\ldots,N\,.
$
The obtained tessellation is then called Centroidal Voronoi Tessellation (CVT) if $\mathbf{p}_i=\mathbf{p}_i^*, \forall i =1,\ldots,N$, where $\mathbf{p}_i^*$ is the centroid of the $i$-th region.

An important property of the CVT is that it represents a solution of the optimization problem:
\begin{equation}
    \min_{\mathbf{p}\in\Omega}\, \sum_i J_i(\mathbf{p}) \quad \text{where} \quad J_i(\mathbf{p})=\int_{V_i}||\mathbf{u}-\mathbf{p}||^2\,d\mathbf{u}
\end{equation}
proving that the Voronoi-based partition allows a good dispersion of the robots over the environment optimizing its coverage. In our context, this ensures having the different signal field measurements taken from well-spread locations, which is crucial to obtain a reliable gradient estimation.

\begin{figure}[tb]
\centering
\includegraphics[width=.75\columnwidth]{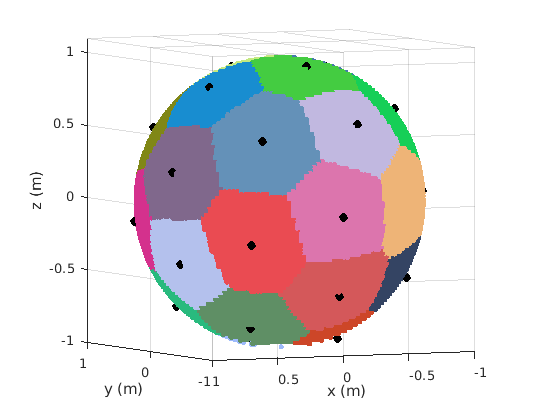}
\caption{Constrained CVT on a spherical surface with 30 generator points.}
\label{fig:cvt}
\end{figure}

A commonly adopted solution to obtain a CVT starting from an arbitrary set of generators is the Lloyd's algorithm \cite{lloyd1982}, whose main steps are: calculate the Voronoi partitions and the relative centers of mass corresponding to the current generators configuration; move each generator to the center of mass of its region; repeat this procedure until the convergence or the maximum number of iterations are reached. Although obtaining Voronoi regions, and consequently CVTs, is quite straightforward in 2D, this is not the case for general surfaces in 3D \cite{du2003voronoi}. To obtain a solution for our case, we firstly discretize our environment by creating a regular mesh of the spherical surface, following the procedure presented in \cite{mesh_sphere}\footnote{Note that this approach does not work for any number of points and is not particularly suitable for small numbers of them.}, to easily compute the Voronoi regions at each iteration. Then, a non-trivial step of implementing the Lloyd's algorithm for a surface in 3D is the computation of the constrained mass centroid $\mathbf{p}_i^c$ of each region $i$. The classical mass centroids $\mathbf{p}_i^*$ are indeed always inside the sphere and a projection on the surface is needed before moving each generator to its new position. However, for a spherical surface, it can be shown that $\mathbf{p}_i^c$ is simply represented by the intersection of the radial line passing through $\mathbf{p}_i^*$ with the spherical surface \cite{du2003voronoi}. Fig.~\ref{fig:cvt} shows an instance of constrained CVT with $30$ generators.

Finally, it is worth noticing that, as shown in \cite{cortes2004coverage}, the Lloyd's algorithm can be implemented in an asynchronous distributed fashion. In this same work, the authors also presented an algorithm to adjust the sensing radius of every robot to ensure that each of them is able to communicate with all its Voronoi neighbors. 
Although the distribution of the complete approach, i.e., including the gradient estimation, is out of the scope of this paper, this is always a desirable property in presence of a multi-robot system. 
In our scenario, this would mean that the robots may find and maintain their positions in the formation in a distributed way only based on the information shared with their teammates.

Following the concepts presented previously, we propose a CVT-based Spherical formation to uniformly distribute a group of robots in a 3D formation. To build such a formation, the robots implement a constrained Lloyd's algorithm on the surface of a sphere. Let us define $\mathbf{p}_i^c$ for $i=1,...,N$ the mass centroids corresponding to a constrained CVT on a sphere centered at the origin and of unitary radius. We express the position $\mathbf{r}_i$ of robot $i$ in the CVT-based Spherical formation with respect to the global frame as:
\vspace{-0.05cm}
\begin{equation}\label{eq:Voronoi_formation}
    \pos_i=\C+D \, \textbf{p}^c_{i}, \quad \forall i=1, \ldots, N
\end{equation}
where $\C$ and $D$ are the center and the radius of the formation.

All the robots in the CVT-based Spherical formation are placed at the same distance from the center, i.e., $\|\pos_i-\C\|=D,\, \forall i$. Their positions, however, are not defined to satisfy by construction the properties in \eqref{eq:Symmetric_properties}. Nevertheless, we can study these properties numerically by considering the robots positions obtained via the constrained Lloyd's algorithm. Let define the formation parameters $\bar{\pos} \in \mathbb{R}^3$ and $\mathbf{M} \in \mathbb{R}^{3\times3}$ as
$$
\bar{\pos}:=\sum_{i=1}^N(\pos_i-\C) \quad \mbox{and} \quad \mathbf{M} := \sum_{i=1}^N(\pos_i-\C)(\pos_i-\C)^T \,.
$$ 
Via numerical analysis, it can be verified that the summation of the relative position vectors, $\bar{\pos}$, remains close to zero\footnote{For the formations used in this work, with a sphere of radius $D=1$, $\bar{\pos} \approx 10^{-4} \mathbf{1}$ where $\mathbf{1}$ denotes the unit vector.} and that the values of matrix $\mathbf{M}$ are close to those of matrix $\frac{ND^2}{3}\textbf{I}$. These properties will be used to study the bound of the gradient estimation error presented in the next section.

\section{3D Cooperative source-seeking}
\label{sec:grad_estimation}

In this section, we present and analyze the cooperative gradient estimation of a 3D signal obtained with Voronoi-based multi-robot formations.  
Although the result on gradient estimation presented in Theorem~\ref{th:Gradient_symmetric} could be still used with a CVT-based formation, the estimation error would be greater due to the geometrical formation properties that are not the same as for the Symmetric Cylindrical formation. To address this issue, we propose a new expression for the gradient estimate by adding two correction terms to obtain a more accurate estimation while keeping the elegant idea of using only the summation of the relative position vectors pondered with the signal measurements and simple computations.

Consider the CVT-based Spherical formation of robots given by \eqref{eq:Voronoi_formation}. The following theorem is proposed:
\begin{Th}\label{th:Gradient_Voronoi}
Assume that $\sigma(\pos):\mathbb{R}^3\rightarrow\mathbb{R}^+$ satisfies Assumption~\ref{assum:signal3D}. Consider a team of $N$ robots forming a CVT-based Spherical formation with radius $D$ centered at $\C$ given by \eqref{eq:Voronoi_formation} and defining the estimated gradient as follows
\vspace{-.0cm}
\begin{equation}\label{eq:Gradient_estimation_Voronoi}
\widehat \nabla \sigma(\C)=\textbf{M}^{-1} \left(\sum_{i=1}^N\sigma(\pos_i)(\pos_i-\C) -\sigma(\C)\sum_{i=1}^N(\pos_i-\C)\right) 
\end{equation}
then it holds
\vspace{-.1cm}
\begin{equation}\label{eq:Gradient_estimation_Voronoi_bound}
\| \widehat \nabla \sigma(\C)-\nabla \sigma(\C) \|\leq \|\textbf{M}^{-1}\|NLD^3 = \|\frac{ND^2}{3}\textbf{M}^{-1}\| \, 3LD.
\end{equation}
\end{Th}

\vspace{.1cm}

\begin{proof}
Using the first-order Taylor expansion of each measurement $\sigma(\pos_i)$ about the point $\C$ and recalling that $\|\widetilde{\pos}_i\|=D$ where $\widetilde{\pos}_i:=\pos_i-\C=D \, \textbf{p}^c_{i}$ as defined in \eqref{eq:Voronoi_formation}, then the following equation holds for all $i=1,\ldots,N$:
\begin{equation}
\nonumber \sigma(\pos_i)-\sigma(\C)=\nabla\sigma(\C)^T\widetilde{\pos}_i+\varphi(\pos_i,\C),
\end{equation}
where $\varphi(\pos_i,\C)$ denotes the remainder of the Taylor expansion. Multiplying the previous equation by $\widetilde{\pos}_i$ and summing over $i=1,\ldots,N$, we obtain
\vspace{-.0cm}
\begin{equation}
\begin{split}
\nonumber \sum_{i=1}^N\sigma(\pos_i)\,\widetilde{\pos}_i-\sigma(\C)\sum_{i=1}^N\widetilde{\pos}_i&=\sum_{i=1}^N\nabla\sigma(\C)^T\,\widetilde{\pos}_i\,\widetilde{\pos}_i+\sum_{i=1}^N\varphi(\pos_i,\C)\,\widetilde{\pos}_i\\
&=\left(\sum_{i=1}^N\widetilde{\pos}_i\,\widetilde{\pos}_i^T\right) \nabla\sigma(\C)+\Phi(D,\C),
\end{split}
\end{equation}
where $\Phi(D,\C)=\sum_{i=1}^N\varphi(\pos_i,\C)\widetilde{\pos}_i$. By using the definition of $\mathbf{M}$, the previous equation can be rewritten as
\vspace{-.1cm}
\begin{equation}
\nonumber \sum_{i=1}^N\sigma(\pos_i)\widetilde{\pos}_i -\sigma(\C)\sum_{i=1}^N\widetilde{\pos}_i
=\textbf{M}\nabla\sigma(\C)+\Phi(D,\C)
\end{equation}
and therefore the gradient estimation \eqref{eq:Gradient_estimation_Voronoi} satisfies
\begin{equation}
\nonumber \widehat \nabla \sigma(\C)
=\nabla\sigma(\C)+\textbf{M}^{-1}\Phi(D,\C).
\end{equation}
According to Assumption~\ref{assum:signal3D} the term $\textbf{M}^{-1}\Phi(D,\C)$ satisfies
\vspace{-.1cm}
\begin{equation}
\nonumber \|\textbf{M}^{-1}\Phi(D,\C)\| \leq \|\textbf{M}^{-1}\| \sum_{i=1}^N L\|\pos_i-\C\|^3 \leq  \|\textbf{M}^{-1}\| NLD^3.\vspace{-.5cm}
\end{equation}
\end{proof}

\begin{Remark}
The previous result relies on the signal measurement at the center of the formation $\sigma(\C)$. Note that, this value can be easily estimated by a simple interpolation computing the average of the measurements collected by the robots, i.e., $\sigma(\C) \approx \hat{\sigma}(\C)=\frac{1}{N}\sum_i^N\sigma(\pos_i)$.
\end{Remark}

This new result allows estimating the gradient of the signal at the center of CVT-based Spherical formations and proves that the estimation error is bounded. The bound of the error is similar to the bound obtained in Theorem~\ref{th:Gradient_symmetric} but multiplied by an additional value that depends on $\textbf{M}$. As previously stated, for a CVT-based Spherical formation, this matrix is close to $\frac{ND^2}{3}\textbf{I}$. Via numerical analysis\footnote{For the formations used in this work, $\| \frac{ND^2}{3}\textbf{M}^{-1}\| \leq 1.04$.}, it can be shown that $\| \frac{ND^2}{3}\textbf{M}^{-1}\| \approx 1$. Consequently, the bound of the estimation error for both results is very similar.

\subsection{Estimation error analysis}
Consider a CVT-based Spherical formation given by \eqref{eq:Voronoi_formation}. Let us now consider that the multi-robot formation computes the gradient estimation by using \eqref{eq:Gradient_estimation}. Following similar derivations as in the proof of Theorem~\ref{th:Gradient_Voronoi}, the gradient estimate \eqref{eq:Gradient_estimation} satisfies
\begin{equation}
\begin{split}
\nonumber \widehat \nabla \sigma(\C) &= \frac{3\sigma(\C)}{ND^2}\sum_{i=1}^N\widetilde{\pos}_i + \frac{3}{ND^2} \mathbf{M} \nabla\sigma(\C)+\frac{3}{ND^2}\Phi(D,\C).
\end{split}
\end{equation}
Let us assume that $\mathbf{M} = \frac{ND^2}{3}\mathbf{I}+\mathbf{M}_\alpha$ where $\mathbf{M}_\alpha$ is a small square matrix\footnote{For the formations used in this work, $\|\mathbf{M}_\alpha\|\leq 0.3$ with a radius $D=1$.}. The estimation error can be expressed as
\begin{equation}
\begin{split}
\nonumber \widehat \nabla \sigma(\C)-\nabla \sigma(\C) =&  \frac{3\sigma(\C)}{ND^2}\bar{\pos} + \frac{3}{ND^2} \mathbf{M}_\alpha \nabla\sigma(\C)+ \frac{3}{ND^2}\Phi(D,\C)
\end{split}
\end{equation}
and the following inequality holds
\begin{equation}
\begin{split}
\nonumber \| \widehat \nabla \sigma(\C)-\nabla \sigma(\C) \|  \leq & \, \frac{3\sigma(\C)}{ND^2} \| \bar{\pos} \| + \frac{3}{ND^2} \| \mathbf{M}_\alpha \| \| \nabla\sigma(\C) \| \\
& + \frac{3}{ND^2}\|\Phi(D,\C)\| \leq \, 3LD + \Phi_\alpha
\end{split}
\end{equation}
where $\Phi_\alpha=\frac{3}{ND^2} \left(\sigma(\C) \| \bar{\pos} \| + \| \mathbf{M}_\alpha \| \| \nabla\sigma(\C) \| \right)$ denotes the additional term which depends on the CVT-based formation properties and on the signal distribution.

We can notice that, as expected, $\Phi_\alpha$ decreases with both the number of robots $N$ and the radius $D$. We then recall that the terms $\sigma(\C)$ and $\| \nabla\sigma(\C) \|$ are both bounded while $\| \bar{\pos} \|$ and $\| \mathbf{M}_\alpha \|$ are close to zero. Moreover, if we focus on the proximity of the signal source, we can see that the term $\| \nabla\sigma(\C) \| \| \mathbf{M}_\alpha \|$ rapidly goes to zero because $\nabla\sigma(\C) \rightarrow 0$ when $\C \rightarrow \pos^*$. However, for strong signals, $\sigma(\C)$ can be a large value and, even if $\|\bar{\pos}\|$ is close to zero, this might lead to a value of $\Phi_\alpha$ that remains larger than the bound of the estimation error obtained in Theorem~\ref{th:Gradient_Voronoi}.

\subsection{Noise analysis}\label{sec:noise}

To consider more realistic scenarios, we assume that the signal measurements collected by the sensors are corrupted by Gaussian zero-mean white noise, i.e.:
\vspace{-.0cm}
$$
y(\pos_i) = \sigma(\pos_i)+\nu_i(\pos_i), \ \ \ \nu_i(\pos_i)\sim\mathcal{N}(0,\nu^2) 
$$
where $y(\pos_i)$ is now the noisy measurement collected by robot $i$ and $\nu^2$ represents the variance of the noise. This situation can be the result of the noise produced by the physical limitations of the sensors and their electronic components. The noise can also model possible small-scale spatial variations of the robots' positions due to turbulence or local perturbations. Let us assume that the noise is independent in each robot measurement, i.e., $\mathbb{E}[\nu_i(\pos_i)\nu_j(\pos_j)]=0, i \neq j$, which is a realistic assumption if $\|\pos_i-\pos_j\|\!>\! D_{min}$ where $D_{min}$ represents the spatial correlation distance of the disturbance.

Since the noise is additive in the measurement, there will be an additional additive estimation error when computing the gradient estimate defined in \eqref{eq:Gradient_estimation}. This additional error caused by the noisy measurements has zero mean and it has been shown that its expected standard deviation is a monotonically decreasing function of the formation radius and the number of robots (see \cite{Lara2019TRO}). 

Another advantage of CVT-based formations is that the minimum distance between two robots is larger than for Symmetrical Cylindrical ones and this difference increases by increasing the number of robots. To illustrate this difference, let us study the minimum distance $d_{min}=min\{\|\pos_i-\pos_j\|, \, \forall i, j, \, i\neq j\}$ for the two formations. For the Symmetric Cylindrical formation, $d_{min}$ can be analytically obtained by simple trigonometric computations. Considering the formation \eqref{eq:Symmetric_formation}, we have $d_{min}=2D\cos(\theta_F)\sin(\frac{2\pi}{N})$. For the CVT-based formation, we obtain this minimum distance via numerical computations. Table~\ref{tab:min_distance} presents the values of $d_{min}$ for the two formations with $D=1$ and different values of $N$.

\begin{table}[ht]
    \centering
    \begin{tabular}{|c|c|c|c|c|}
    \hline
       $N$  & 6 & 8 & 10 & 20 \\
       \hline
        Symmetric Formation $d_{min}$ & 1.00 & 0.81 &  0.68 & 0.37\\
        CVT-based Formation $d_{min}$ & 1.41 & 1.14 & 1.06 & 0.76\\
    \hline
    \end{tabular}
    \caption{Minimum distance $d_{min}$ with respect to $N$. \vspace{-0.6cm}}
    \label{tab:min_distance}
\end{table}

This property is important because it strengthens the assumption about the independence of noisy measurements and it proves a better uniform distribution of the robots around the center of the formation.
As shown in the next section, this leads to an increased robustness to asymmetric noisy measurements, for example, when one robot sensor has a level of noise greater than the rest of the team.

\begin{figure}[t]
     \centering
    \includegraphics[trim=1.5cm 0cm 2.5cm 0cm, clip=true, width=.9\columnwidth]{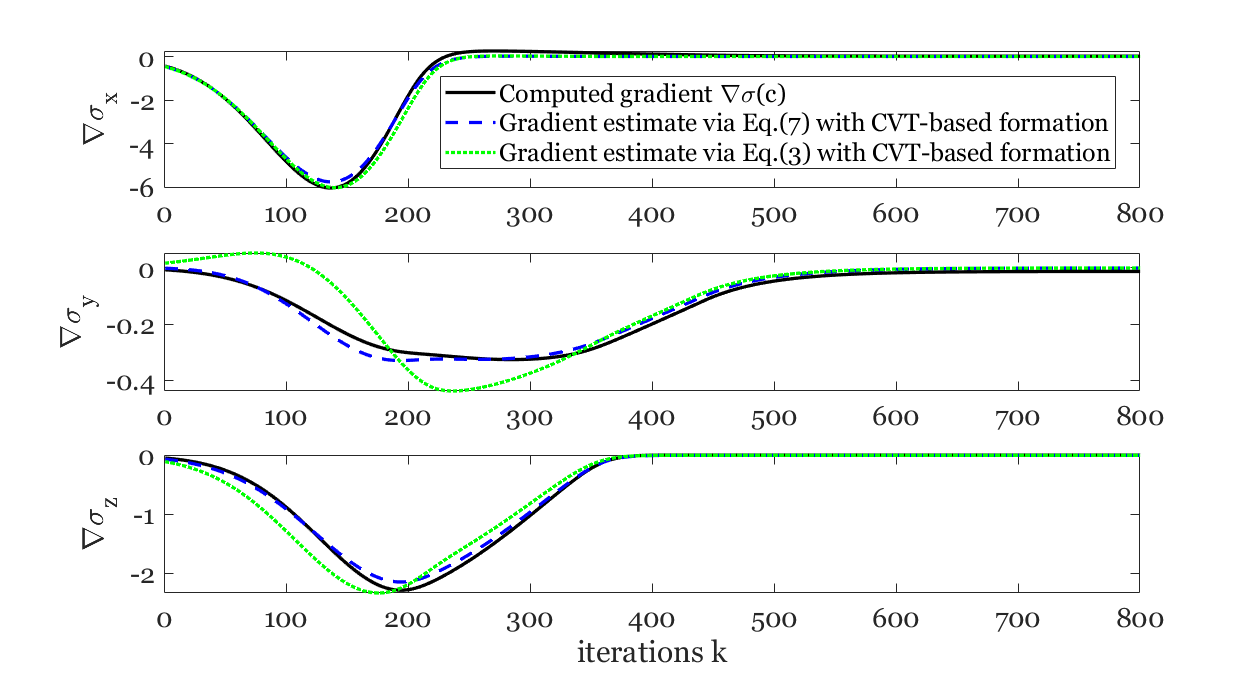}
        \caption{Evolution of the three components of the gradient estimate $\widehat \nabla\sigma(\C)$ for a team of $N=7$ robots with CVT-based Spherical formations compared to the real computed gradient of $\sigma_1(\pos)$.} \vspace{-.2cm}
        \label{fig:GA_comparison}
\end{figure}

\section{Simulation Results and Analysis}
\label{sec:results}

In this section, we present numerical simulations to illustrate the performance of the proposed cooperative estimation algorithms applied to the source-seeking problem in 3D. The signals considered in the simulations are Gaussian functions whose surface sets are ellipsoids given by
\vspace{-.0cm}
$
\sigma_l(\pos)=A_le^{-\pos^T\textbf{S}_l\pos}
$
where $A_l$ determines the intensity of the signal strength and $\textbf{S}_l \in \mathbb{R}^{3\times3}$ is a symmetric matrix. The maximum corresponding to the source is located at $\pos^*\!=\![0, 0, 0]^T$.

In our analysis, we study two different scenarios: first the case of a strong signal without any noise and a second case with a weaker signal where measurements are affected by noise. The signal functions for these two scenarios are: $\sigma_1(\pos)$ with $A_1=100$ and $\textbf{S}_1\!=\!10^{-4}\left[\begin{smallmatrix}
100 & 1 & 1\\
1 & 1 & 0\\
1 & 0 & 10\\ 
\end{smallmatrix}
\right]$ and $\sigma_2(\pos)$ with $A_2=1$ and $\textbf{S}_2\!=\!10^{-4}\left[\begin{smallmatrix}
20 & 5 & 2\\
5 & 20 & 2\\
2 & 2 & 100\\ 
\end{smallmatrix}
\right]$ respectively. For all the results, simulation parameters are set to $\varepsilon=1$ and $\gamma=0.1$. 

Fig.~\ref{fig:source-seeking} illustrates a CVT-based formation with $D=4$ and an odd number of robots ($N=7$) performing a GA algorithm where the gradient of $\sigma_1$ is estimated by \eqref{eq:Gradient_estimation_Voronoi}. The red line represents the trajectory of the formation center, the blue circles represent the initial robots' positions and the green ones an intermediate state at iteration $k=500$. The gradient estimation for this scenario is shown in Fig.~\ref{fig:GA_comparison}. The exact computation of the gradient $\nabla\sigma(\C)$ is compared with two gradient estimates $\widehat{\nabla}\sigma(\C)$ computed by \eqref{eq:Gradient_estimation} and by \eqref{eq:Gradient_estimation_Voronoi} respectively with the same CVT-based formation. The data show how the correction term presented in \eqref{eq:Gradient_estimation_Voronoi} provides an estimation closer to the exact gradient.

\begin{figure}[t]
     \centering
    \includegraphics[trim=1.5cm 0cm 2.3cm 0cm, clip=true, width=.85\columnwidth]{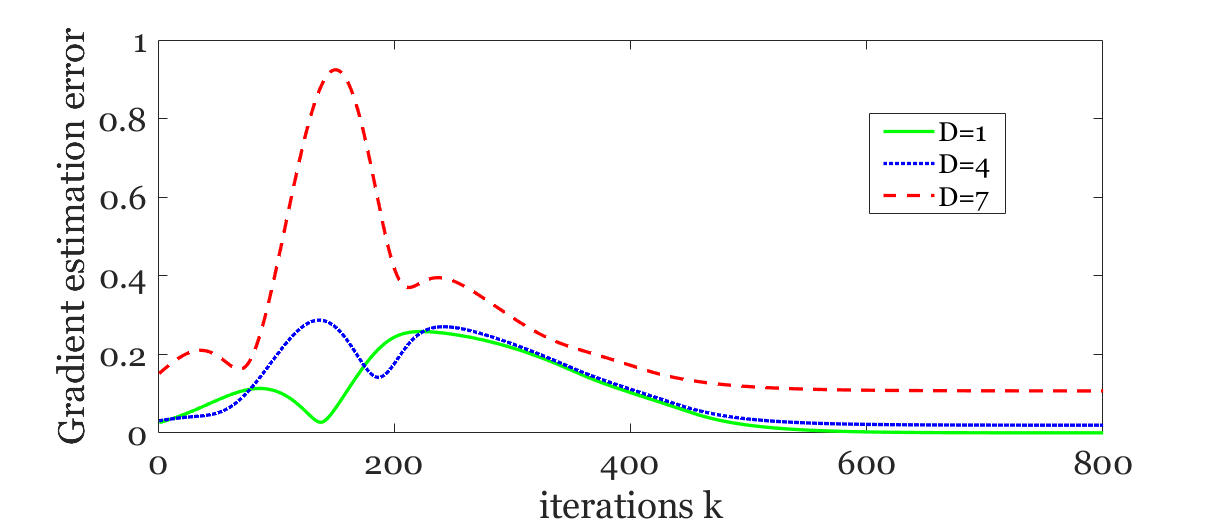}
        \caption{Evolution of the norm of the gradient estimation error, i.e., $\|\widehat\nabla\sigma(\C)\-\nabla\sigma(\C)|$, with a CVT-based Spherical formation for different values of the formation radius $D$. \vspace{-.2cm}}
        \label{fig:varying_D}
\end{figure}

We then analyze the effect of the radius $D$ on the gradient estimation error. Fig. \ref{fig:varying_D} shows the evolution of the norm of this error, i.e., $\|\widehat\nabla\sigma(\C)-\nabla\sigma(\C)\|$, where $\widehat\nabla\sigma(\C)$ is computed by \eqref{eq:Gradient_estimation_Voronoi}, for three different values of $D$. This study supports the analytical results given in \eqref{eq:Gradient_estimation_Voronoi_bound}: as expected the norm of the error increases monotonically with the radius. In particular, we can see that, although $D=4$ remains significantly closer to $D=1$ than $D=7$, its final error at convergence is larger and the minimum error is obtained with $D=1$.

\begin{figure}[t]
     \centering
     \includegraphics[trim=1.8cm 0cm 2.5cm 0cm, clip=true, width=.82\columnwidth]{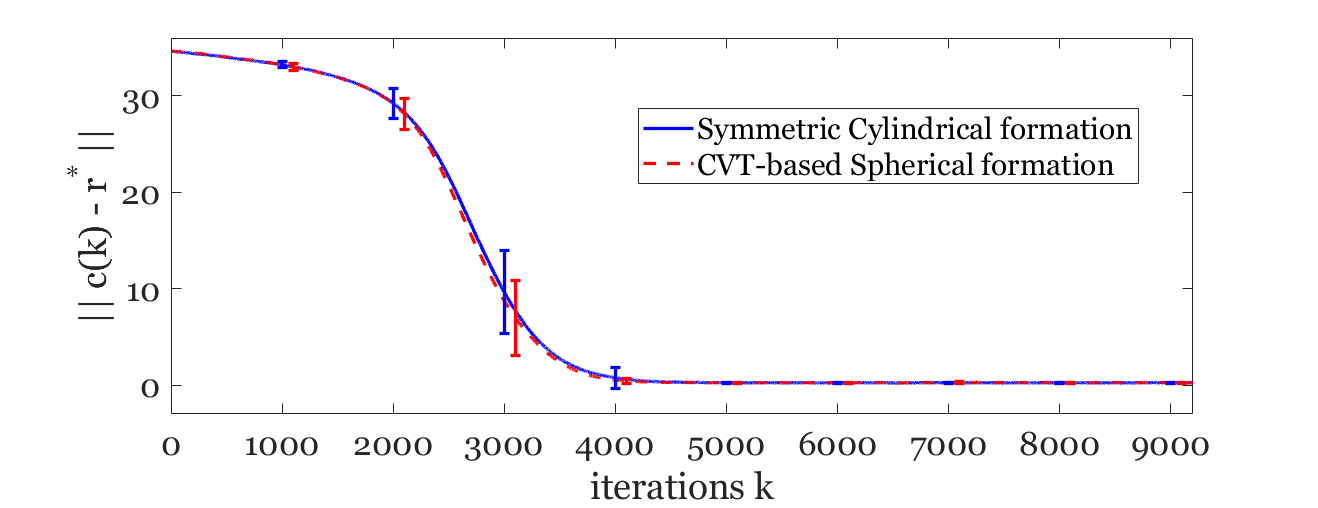} \\
    \includegraphics[trim=1.8cm 0cm 2.5cm 0cm, clip=true, width=.82\columnwidth]{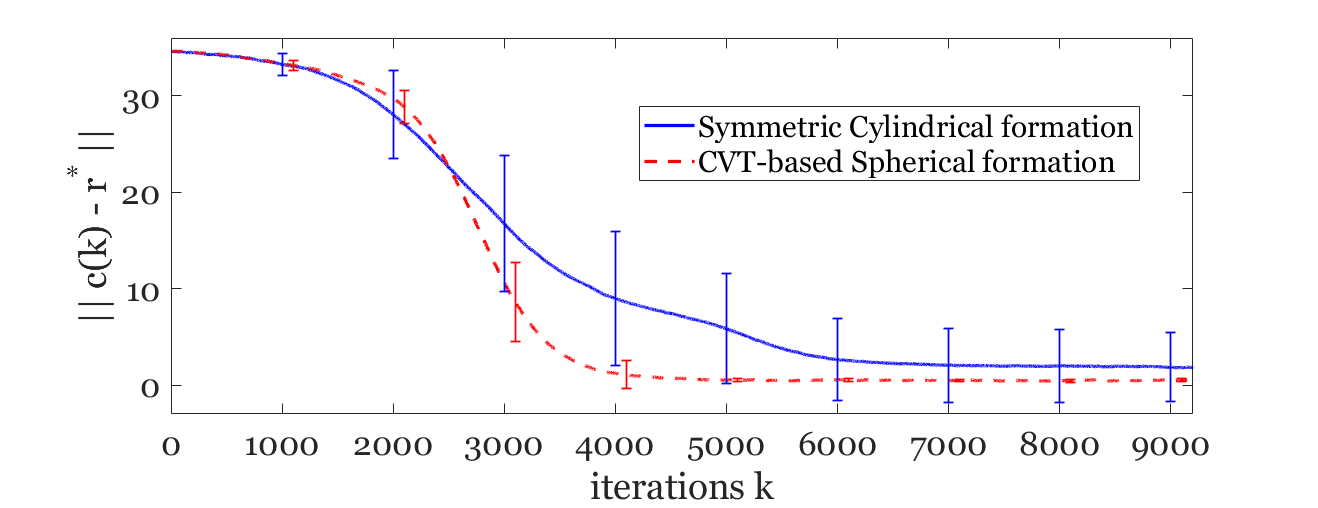}
        \caption{Mean and standard deviation over $100$ trials of the distance between the center and the source location for both Symmetric and CVT-based formations in presence of noisy measurements with $\nu = 0.1$ for all robots (top) and with $\nu = 0.5$ only for one robot and $\nu = 0.1$ for the others (bottom).\vspace{-.2cm}}
        \label{fig:source-seeking_noise}
\end{figure}

Let us now consider the second scenario of a weaker signal with robots' measurements affected by noise. To show the robustness of our solution to noisy measurements, we compare the obtained results with the Symmetric Cylindrical formation presented in \cite{Lara2019TRO}. Note that comparisons with other alternative solutions are difficult to obtain due to the very limited attention that the current literature has dedicated to the case of 3D fields with noisy signal measurements (see discussion in Section \ref{sec:intro}). Fig.~\ref{fig:source-seeking_noise} shows a statistical analysis of the evolution of the distance $\|\C-\pos^*\|$ for a series of $100$ trials to compare the influence of noisy measurements on the source-seeking convergence for a Symmetric Cylindrical formation and a CVT-based Spherical formation of $N=8$ robots. The top figure shows the results when $\nu=0.1$ for all the robots in the formation. In this situation, the behavior of the two formations is almost identical. The figure at the bottom shows the results where one robot obtains faulty measurements, modeled by a higher level of noise $\nu=0.5$, while for all the others $\nu=0.1$. This asymmetry in the noise level has an important influence on the gradient estimation when computed by the Symmetric Cylindrical formation that strongly relies on its geometrical properties to ensure a good gradient estimate. Consequently, the convergence toward the source is affected resulting in a greater final error and a much larger standard deviation along its entire trajectory. On the other hand, the CVT-based formation is less affected and presents a behavior not far from the equal noise case. This represents a significant improvement in robustness against unexpected faulty measurements that can be of paramount importance in real applications.

\section{Conclusion}
\label{sec:conclusion}
This paper presented a new strategy to design suitable multi-robot formations that allow an accurate and robust estimation of the gradient of a 3D signal field generated by a source of unknown position. This estimation, obtained via simple analytic expressions that ensure a low computation cost, is then leveraged to guide the multi-robot formation through the field to finally localize the source. Results in simulations supported the analytic analysis of the proposed approach. Moreover, they showed how it is able to outperform an alternative approach, providing a higher robustness with respect to faulty measurements.

For future work, our primary intent is to build upon these results to obtain the full distribution of this solution, including limited communication capabilities, while maintaining theoretical guarantees on the gradient estimation.

\bibliographystyle{IEEEtran}
\bibliography{biblio}

\end{document}